\documentclass[sigconf]{acmart}
\AtBeginDocument{%
  }

\copyrightyear{2025}
\acmYear{2025}
\setcopyright{cc}
\setcctype{by}
\acmConference[WWW Companion '25]{Companion Proceedings of the ACM Web Conference 2025}{April 28-May 2, 2025}{Sydney, NSW, Australia}
\acmBooktitle{Companion Proceedings of the ACM Web Conference 2025 (WWW Companion '25), April 28-May 2, 2025, Sydney, NSW, Australia}
\acmDOI{10.1145/3701716.3715472}
\acmISBN{979-8-4007-1331-6/2025/04}

\begin{document}

\title{Navigating Semantic Relations: Challenges for Language Models in Abstract Common-Sense Reasoning}

\author{Cole Gawin}
\email{gawin@usc.edu}
\affiliation{%
  \institution{University of Southern California}
  \city{Los Angeles}
  \state{California}
  \country{USA}
}

\author{Yidan Sun}
\email{yidans@isi.edu}
\affiliation{%
  \institution{University of Southern California}
  \city{Los Angeles}
  \state{California}
  \country{USA}
}

\author{Mayank Kejriwal}
\email{kejriwal@isi.edu}
\affiliation{%
  \institution{University of Southern California}
  \city{Los Angeles}
  \state{California}
  \country{USA}
}

\renewcommand{\shortauthors}{Cole Gawin, Yidan Sun, and Mayank Kejriwal}

\begin{abstract}
 Large language models (LLMs) have achieved remarkable performance in generating human-like text and solving reasoning tasks of moderate complexity, such as question-answering and mathematical problem-solving. However, their capabilities in tasks requiring deeper cognitive skills, such as common-sense understanding and abstract reasoning, remain under-explored. In this paper, we systematically evaluate abstract common-sense reasoning in LLMs using the ConceptNet knowledge graph. We propose two prompting approaches: \textit{instruct prompting}, where models predict plausible semantic relationships based on provided definitions, and \textit{few-shot prompting}, where models identify relations using examples as guidance. Our experiments with the gpt-4o-mini model show that in instruct prompting, consistent performance is obtained when ranking multiple relations but with substantial decline when the model is restricted to predicting only one relation. In few-shot prompting, the model's accuracy improves significantly when selecting from five relations rather than the full set, although with notable bias toward certain relations. These results suggest significant gaps still, even in commercially used LLMs' abstract common-sense reasoning abilities, compared to human-level understanding. However, the findings also highlight the promise of careful prompt engineering, based on selective retrieval, for obtaining better performance.
\end{abstract}




\maketitle

\section{Background}
Recent advancements in large language models (LLMs) have demonstrated their impressive ability to generate humanlike text and solve reasoning challenges with moderate complexity, ranging from question answering to mathematics \cite{llmCap,MCS4}. However, their performance can lack robustness when subjected to tasks that require deeper cognitive skills, such as common-sense understanding and abstract reasoning \cite{MCS1,MCS2}. These abilities are especially important in applications (such as healthcare and robotics) where requirements and inputs may ambiguously specified, but that require high reliability \cite{app1,app2}. 
\textit{Common-sense} encompasses intuitive knowledge about the world i.e., understanding everyday concepts and their relationships \cite{davis2015commonsense,santos2024theoretically}. \textit{Abstract reasoning} is generally defined as the ability to detect patterns and relationships, extrapolate from examples, and make connections between different ideas and concepts with minimum reliance on memorized knowledge \cite{cattell1971abilities,marini1994development,shen2023experimental}.  

Although distinct, there is a clear overlap between both common-sense and abstract reasoning. However, there remain open challenges in evaluating common-sense abstract reasoning abilities in LLMs systematically and rigorously, especially when both abilities are involved in the same test \cite{eval1}. To minimize noise, such an evaluation requires establishing clear and precise criteria \cite{eval2}, and must be systematic enough to be replicable. In recent years, AI evaluation (especially concerning LLMs) has emerged as a complex research agenda in its own right \cite{AIaudit,2024challenges}, with concerns ranging from data, pre-training, and model contamination \cite{2022designing}, to inappropriate (often, theory-agnostic) design of reasoning benchmarks for assessing generalization \cite{2024can,2021essential}. 

In this paper, we propose and conduct such an evaluation using the ConceptNet knowledge graph \cite{speer2017conceptnet}. ConceptNet encapsulates a broad array of general common-sense knowledge about the world. Entries in ConceptNet link two nodes together by edges representing semantic relations; for instance, \texttt{/c/en/car} would be linked to \texttt{/c/en/vehicle} by an edge labeled \texttt{/r/IsA}.\footnote{Note that ConceptNet is a multigraph, and parallel edges with unique labels between nodes are allowed.} ConceptNet contains a mix of words, short phrases, and common sayings, thereby serving as a diverse basis upon which to evaluate the aforementioned cognitive abilities \cite{conceptnet1}. Although the semantic types in ConceptNet are broad and shown to be amenable to further substructure analysis \cite{conceptnet2}, their common-sense nature also facilitate complex evaluations \cite{conceptneteval1,conceptneteval2}.  

Our specific contributions in this paper are as follows. Using ConceptNet as a base, we describe an experimental study using two prompting approaches for evaluating LLMs on common-sense abstract reasoning: \textit{instruct prompting with named relations} and \textit{few-shot prompting with unnamed relations}. The former tests the model's ability to use explicit relation names (like \texttt{/r/IsA}, \texttt{/r/HasA}, and \texttt{/r/MadeOf}) and definitions to predict the correct semantic relation, given a pair of head and tail entities. The latter uses few-shot prompting to assess an LLM's ability to generalize from analogous examples \textit{without} explicit definitions or names, thereby testing the model's implicit understanding of relationship semantics.  

    \begin{figure*}
    \centering
    \includegraphics[width=\textwidth]{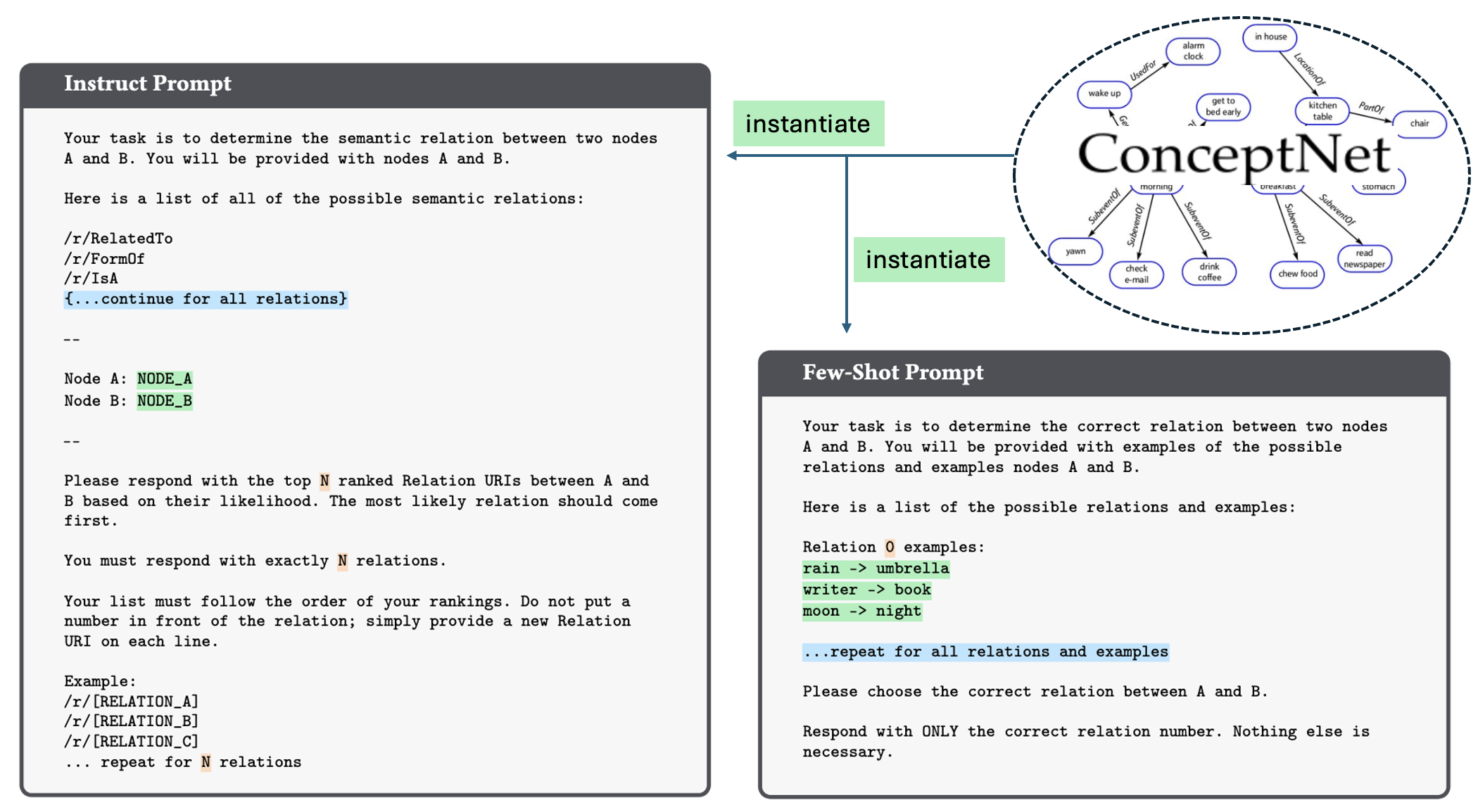} 
    \caption{Schematized representation of two prompting templates/approaches (instruction prompting and few-shot prompting) for evaluating an LLM on abstract common-sense reasoning. The actual prompts given to the model are instantiated using the ConceptNet knowledge graph.}
    \label{figExample}
\end{figure*}

Using gpt-4o-mini as our test model due to its commercial availability and widespread use \cite{yang2024ecosystem}, we systematically evaluate abstract common-sense reasoning through these two prompting approaches. Our experiments assess the model's ability to identify semantic relationships under different experimental conditions. The remainder of this paper details our methodology and experimental findings and discusses future research directions.


\section{Evaluation Framework}

To ensure consistency and reproducibility in our evaluation, we developed two stable datasets: one for evaluation and another for providing the LLM with relation examples (which is important for the few-shot prompting experiments). The evaluation dataset was constructed by randomly sampling edges corresponding to each semantic relation from ConceptNet, subject to specific constraints to maintain the integrity of the samples. First, only edges connecting English nodes ($E_{en}$) were considered. Next, we filtered out reflexive edges, ensuring that the two nodes connected by an edge were distinct ($E_{distinct}$). Finally, we selected unique edges ($E_{unique}$), eliminating duplicates where multiple edges with the same argument nodes ($arg_1,arg_2$) were present. The resulting subset ($R$) provided the foundation for our evaluation, from which 100 samples were randomly drawn for each semantic relation. The second dataset, designed for semantic relation examples, was synthetically generated to provide prototypical representations of each relation. Using the \texttt{gpt-4o-mini} model \cite{achiam2023gpt}, we generated three examples for each semantic relation in the format 
$NODE_A \rightarrow NODE_B$. These examples were then parsed and stored for consistency. Unlike the evaluation dataset, which relied on random sampling from ConceptNet, this approach ensured that the examples were clear, representative, and directly aligned with the semantic relation under consideration. Next, we describe the two specific prompting approaches designed to evaluate the effectiveness of LLMs in predicting common-sense semantic relations between pairs of entities (Figure \ref{figExample}). All experiments were conducted using \texttt{gpt-4o-mini} via the OpenAI API.






\textbf{Instruct Prompting with Named Relations.} The first methodology involves providing the model with exact definitions for each semantic relation. The model is asked to respond with the top $N$ ranked relations for each pair of nodes. Relations are explicitly named using their actual identifiers such as \texttt{IsA}, \texttt{HasA}, and \texttt{MadeOf}. The prompt for this methodology includes a task description that instructs the model to determine the correct relation between two nodes. It also contains a comprehensive list of all possible relations along with their precise definitions, and the specific nodes A and B for which the relation is to be determined.










The predicted rankings are then evaluated using Normalized Discounted Cumulative Gain (NDCG) \cite{jarvelin2002cumulated}, a metric that measures the quality of the ranking provided by the model. The NDCG metric is calculated by first computing the Discounted Cumulative Gain (DCG) for a particular ranked list ($\text{DCG}_p = \sum_{i=1}^p \frac{2^{\text{rel}_i} - 1}{\log_2(i + 1)}$), 
where \( \text{rel}_i \) is the relevance score of the result at position \( i \) and \( p \) is the number of results in the list. The ideal DCG (IDCG) is the DCG value of the ideal ranking. The NDCG at position \( p \) is then computed as $\text{NDCG}_p = \frac{\text{DCG}_p}{\text{IDCG}_p}$.


In the context of our experiment, the relevance score (\( \text{rel}_i \)) is binary, indicating whether the predicted relation is correct (1) or incorrect (0).
Using NDCG allows for a finer-grained evaluation of the model's performance, rather than a simple accuracy evaluation ($N=1$) of whether the model predicted the correct relation. We performed the instruct prompting methodology on four separate variations, wherein the model was asked to respond with the top 10, 5, 3, and 1 ranked relation(s) for the provided pair of nodes. We calculated the NDCG score for each individual pair presented to the LLM and also calculated the average NDCG score for each iteration.

\textbf{Few-Shot Prompting with Unnamed Relations.} In the second methodology, the model was provided with examples for each relation type. The model's task in this methodology is to identify the correct relation using only the provided examples as guidance, as opposed to using explicitly provided definitions or names. The evaluation focuses on whether the correct relation can be identified by the model from among the shuffled list of relations. Accuracy is assessed by comparing the model’s predicted relation with the actual correct relation.









Instead of using the actual names of the relations within the prompt, the relations were labeled as Relation 1, Relation 2, and so on. The prompt includes a task description, several examples demonstrating the nature of each relation, and the specific nodes A and B to be evaluated. A mix of the correct relation and randomly selected other relations are included in each prompt to challenge the model's understanding. Because of the manner in which this prompting is designed, it serves as a bulwark against LLMs merely returning triples that they might have memorized (e.g., if ConceptNet, which is publicly available, was used in the pretraining of these models). This also makes it a better test of abstract common-sense reasoning than the first methodology. Nevertheless, we use both methodologies for comparative purposes.

We performed the few-shot prompting methodology on two separate variations: one in which the model was allowed to choose from all possible relations, and one in which the model was provided with five possible relations (inclusive of the correct option) to choose from. To evaluate the model's performance in predicting the correct relations, we calculated Cohen's kappa score, a correlation metric that is popularly used in the information retrieval community\cite{doi:10.1177/001316446002000104, 10.1162/coli.07-034-R2}, to compare the predicted relations against the actual relations for both iterations.


\section{Results}

\begin{figure*}
    \centering
    \includegraphics[width=5.8in]{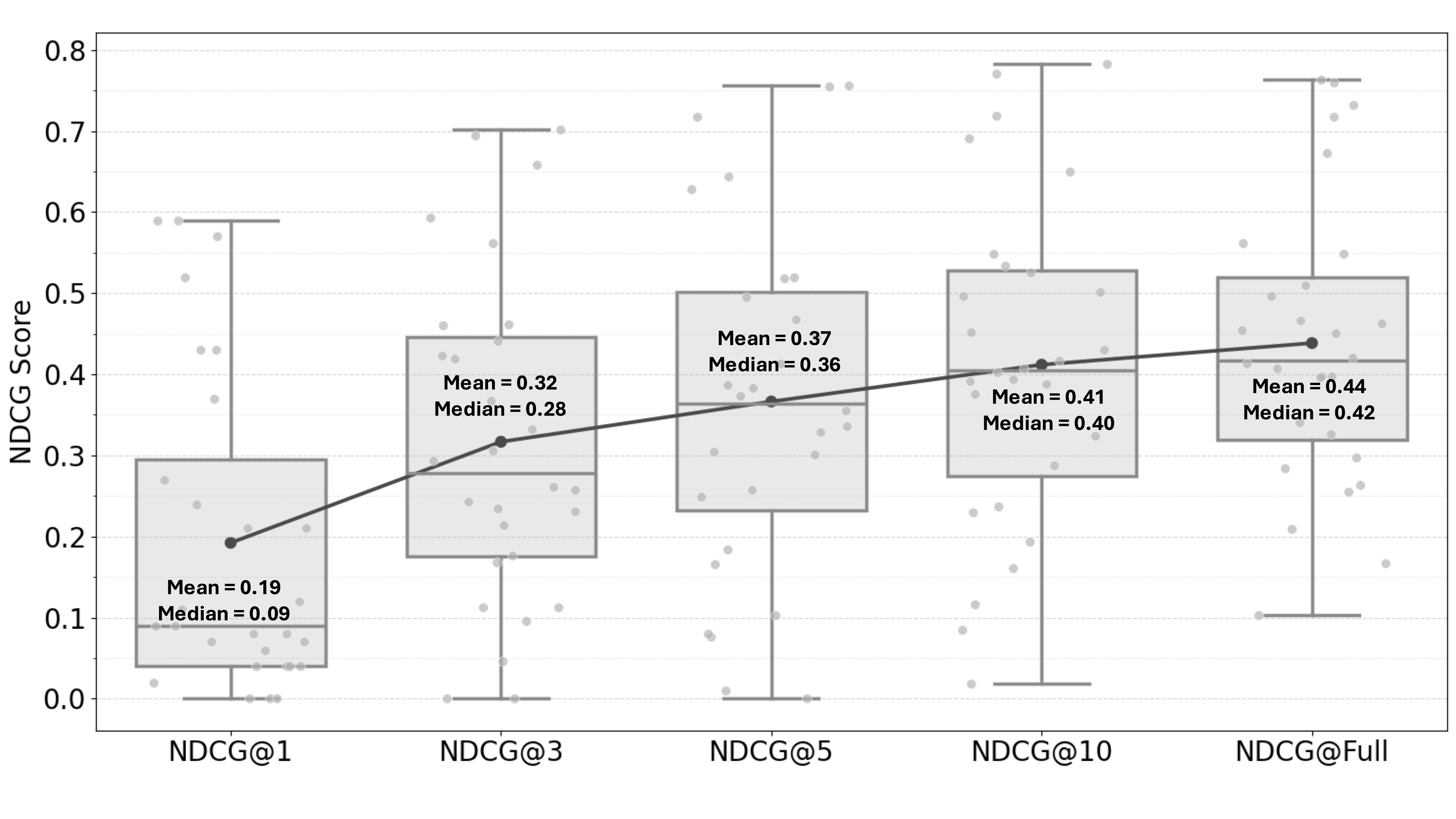} 
    \caption{Distribution of NDCG scores grouped by relations across different K values. Boxplots represent quartiles and ranges for average NDCG scores as the model ranks different numbers of relations (top 10, 5, 3, and 1; shown as NDCG@1, NDCG@3, NDCG@5, NDCG@10, and NDCG@Full). Individual gray points show the average NDCG score for each relation within each K category. Mean NDCG scores are represented by black dots inside each box.}
    \label{fig1}
\end{figure*}

On the \textit{instruct prompting} experiments, we found negligible differences in the average NDCG score as the model was allowed to rank more or less (top 10, 5, 3, and full) relations in its response unless the model was restricted to responding with only one possible relation (Figure \ref{fig1}). 
Notably, for the majority of pairs, the correct relation was not returned in the top 1 across most rankings, showing that even a commercial model like gpt-4o-mini needs significant improvement before it can match human-level common-sense. This is in contrast with performance that has been reported for question-answering benchmarks \cite{MCS1,MCS3}. However, because our benchmark construction is more heavily dependent on semantics, rather than pure natural language, and also because it requires the model to rank (rather than pick from multiple-choice) options, we surmise that our test is more challenging for LLMs.
In additional experiments (not reproduced herein), we also found that the relations with the highest average NDCG scores differed slightly between variations and that there was a small dataset bias: across all iterations, \texttt{/r/UsedFor}, \texttt{/r/PartOf}, and \texttt{/r/Antonym} were predicted correctly within the top 5 relations with the highest average NDCG scores.


\begin{figure*}
    \centering
    \includegraphics[width=\textwidth]{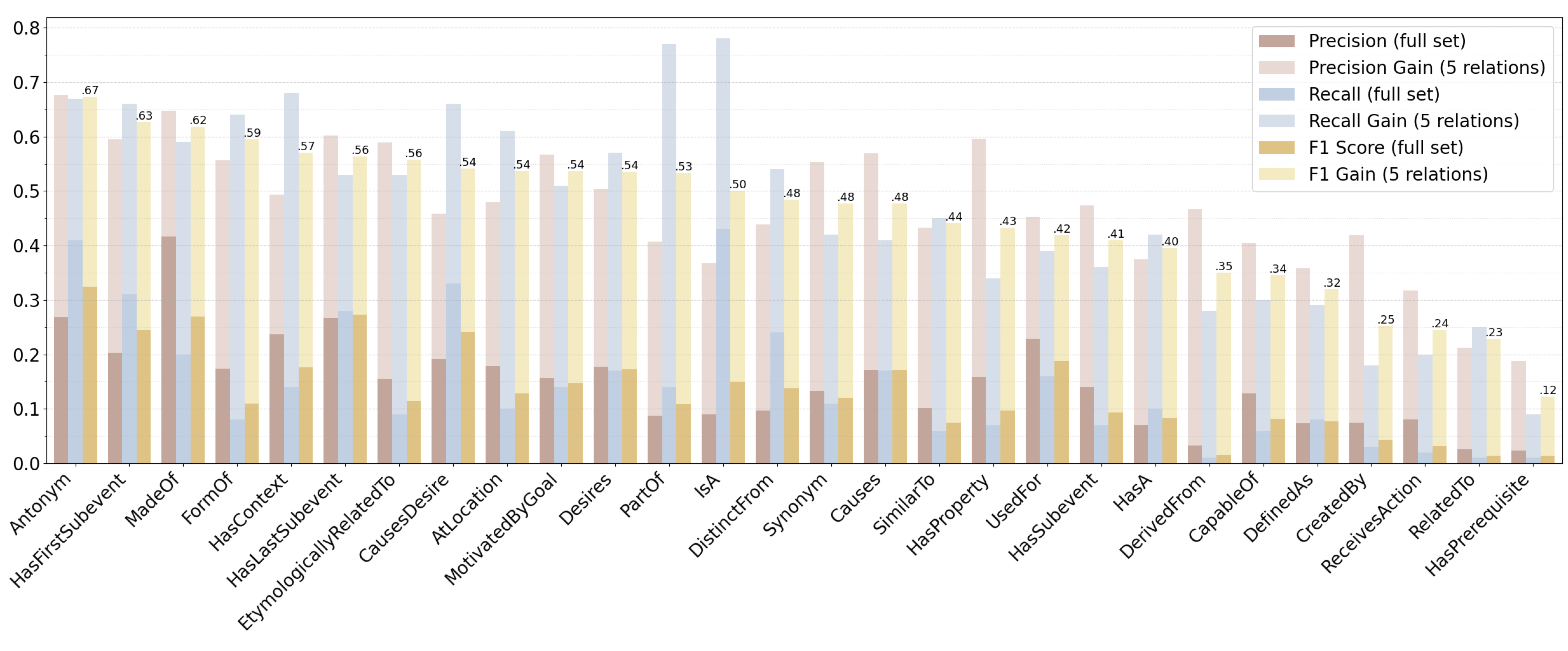} 
    \caption{Comparison between two \textit{few-shot prompting} experimental conditions: when the model was provided with only \textit{five}  (lighter shade) versus \textit{full set} (darker shade) of possible relations. Note that the lighter shaded portion shows the additional performance gain achieved when limiting choices to 5 relations, with the total height of each bar (dark + light portions combined) representing performance in the 5-relation setting. F1 scores for these are explicitly noted above each bar-set.}
    \label{fig2}
\end{figure*}

On the \textit{few-shot prompting} experiments, we observed striking differences in the results for the two separate variations we performed for this experiment (Figure \ref{fig2}).  
We also repeated the analysis using an association-based metric like Cohen's $\kappa$ and similarly found that it was significantly higher ($\kappa=0.385$) when the model was only provided with five possible relations, compared to when the model was provided with \textit{all} possible relations ($\kappa=0.094$).


Some evidence was also found for dataset bias in the few-shot prompting experiment. When allowed to choose from all possible relations, \texttt{/r/IsA} was often the most confused with or mistaken for other categories. Specifically, there were 13 relations for which \texttt{/r/IsA} was (incorrectly) predicted more frequently than the actual relation.
However, this widespread confusion seems to dissipate when the model is only presented with five possible relations to choose from, which suggests a promising next step (restrict, then rank, as opposed to rank all possible relations) for improving the model's abstract common sense reasoning ability. Nevertheless, in these experiments, we manually provided the model with five relations, out of which one was correct. Further experimentation is needed to determine whether the model is capable of restricting the choices to fewer relations, prior to ranking, on its own using techniques inspired by (for example) retrieval augmented generation or RAG \cite{gao2023retrieval}.


\section{Conclusion and future work}

In this paper, we propose and conduct a systematic evaluation of abstract common-sense reasoning abilities in large language models using the ConceptNet knowledge graph. Our study utilizes two prompting methods to assess the capabilities of gpt-4o-mini, including \textit{instruct prompting with named relations} and \textit{few-shot prompting with unnamed relations}. Our results indicate that while gpt-4o-mini shows a foundational understanding of common sense knowledge, it often struggles to accurately identify the correct relations, especially in more abstract reasoning tasks without explicit guidance. Specifically, in our \textit{instruct prompting} experiments, the model maintained consistent NDCG scores when outputting multiple relations, whether selecting from the top 3, 5, 10, or all possible relations. However, its performance declined significantly when restricted to predicting a single relation. In our \textit{few-shot prompting} experiments, the model's accuracy improved substantially when selecting from five possible relations rather than the full set, though we observed notable bias toward certain relations, particularly `/r/IsA'.

An important line of future research is to replicate the experiment on models other than gpt-4o-mini; specifically models like Llama and OpenAI's o1. However, it bears noting that the latter is still cost-prohibitive at the time of writing, similar to gpt-4. For this reason, gpt-4o-mini still continues to be the preferred model in many settings owing to its high performance\footnote{Even in research studies that have used gpt-4o-mini in conjunction with other models, performances are found to be highly correlated \cite{mirzadeh2024gsm}. Thus far, we are not aware of a single study where performance on gpt-4o-mini has stalled or is low, while being high in other advanced models like gpt-4.} (generally in the range of the more advanced models for many tasks \cite{mirzadeh2024gsm}). Nevertheless, replicating the results on other models, and also on other prompts, would provide a more rigorous basis for quantifying abstract common sense in LLMs.    

\textbf{Ethical use of data and informed consent.} N/A

\bibliographystyle{ACM-Reference-Format}
\bibliography{references}

\end{document}